\RequirePackage{fix-cm}
\documentclass{svjour3}                     
\smartqed  

\usepackage{multirow}
\usepackage{epsfig}
\usepackage{algorithm}
\usepackage{algorithmic}
\usepackage{amssymb}
\usepackage{cite}
\usepackage{amsmath}
\usepackage{extarrows}
\usepackage{xcolor}
\usepackage{amsmath}
\usepackage{clrscode}
\usepackage{url}
\usepackage{enumerate}
\usepackage{subfigure}
\usepackage{booktabs}
\usepackage{array}
\usepackage{helvet}
\usepackage{courier}
\usepackage{graphicx}
\usepackage{bm}
\usepackage{makecell} 
\usepackage{diagbox} 
\newtheorem{myDef}{Definition}
\newtheorem{myPro}{Problem}

\begin{document}

\title{MPE: A Mobility Pattern Embedding Model for Predicting Next Locations}

\author{Meng~Chen \and Xiaohui~Yu \and Yang~Liu}

\institute{Meng~Chen \at School of Information Technology, York University, Toronto, Canada \\ \email{mchen16@yorku.ca}  \\
           \and
           Xiaohui~Yu \at School of Information Technology, York University, Toronto, Canada \\ \email{xhyu@yorku.ca}   \\
               \and
           Yang~Liu \at School of Computer Science and Technology, Shandong University, Jinan, China \\ \email{yliu@sdu.edu.cn}   \\
}

\maketitle

\begin{abstract}
The wide spread use of positioning and photographing devices gives rise to a deluge of traffic trajectory data (e.g., vehicle passage records and taxi trajectory data), with each record having at least three attributes: \textit{object ID}, \textit{location ID}, and \textit{time-stamp}. In this paper, we propose a novel mobility pattern embedding model called MPE to shed the light on people's mobility patterns in traffic trajectory data from multiple aspects, including sequential, personal, and temporal factors. MPE has two salient features: (1) it is capable of casting various types of information (object, location and time) to an integrated low-dimensional latent space; (2) it considers the effect of ``phantom transitions'' arising from road networks in traffic trajectory data. This embedding model opens the door to a wide range of applications such as next location prediction and visualization. Experimental results on two real-world datasets show that MPE is effective and outperforms the state-of-the-art methods significantly in a variety of tasks.

\keywords{Human Mobility Patterns \and Embedding Learning \and Traffic Trajectory Data \and Next Location Prediction}
\end{abstract}

\section{Introduction}

\begin{figure}
\centering
\subfigure[Electronic taxi dispatching system.]{
\includegraphics[width=0.4\textwidth]{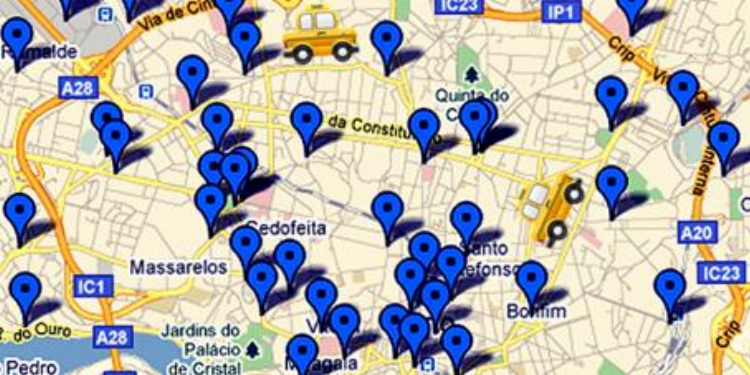}}
\hspace{0.05\textwidth}
\subfigure[Traffic surveillance system.]{
\includegraphics[width=0.4\textwidth]{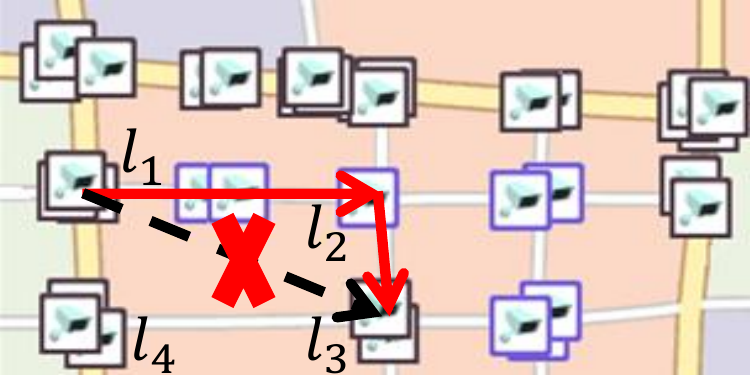}}
\caption{Traffic trajectory data.}
\label{fig:intro}
\end{figure}

The increasing prevalence of electronic dispatch systems and surveillance devices has made it possible to collect a massive amount of traffic trajectory data. For example, as shown in Fig.~\ref{fig:intro}(a), the mobile data terminals installed in each taxi could typically provide information on GPS (Global Positioning System) localization and taximeter state \cite{de2015artificial}. As another example, vehicles are photographed when they pass by the surveillance cameras (as depicted in Fig.~\ref{fig:intro}(b)), and structured vehicle passage records (VPRs) are subsequently extracted from the pictures using optical character recognition (OCR) \cite{chen2015mining,Zhen2014GrandLand}. The data collected in both scenarios contain at least three attributes: \textit{object ID}, \textit{location ID}, and \textit{time-stamp}, which provide an opportunity to deeply understand people's mobility patterns.

While new technologies have made it possible to see where a vehicle has been, it is still non-trivial to predict where it is going next in a real-world transportation system. Among other potential applications, accurate prediction of next locations can help improve the effectiveness of electronic taxi dispatching systems and city-scale traffic management. For example, if the dispatchers know approximately where their taxis will arrive next, they would be able to identify which taxi to assign to each pickup request; if the transportation management system is aware of where the vehicles will go next, it could adjust traffic signal timing dynamically to help relieve traffic congestion.

Despite its great practical value, it is challenging to analyze and mine traffic trajectory data to predict next locations, due in part to the following important but often overlooked considerations:
\begin{enumerate}
\item { \bf Road network constraints}. The routes of vehicles have to follow the road networks. For example, as shown in Fig.~\ref{fig:intro}(b), a vehicle could take a route $ l_1 \rightarrow l_2 \rightarrow l_3$, while it would be impossible to observe that a vehicle moves directly from $l_1$ to $l_3$ without passing $l_2$ or $l_4$, where $l_1$, $l_2$, $l_3$ and $l_4$ are locations. 
\item {\bf Personal tendencies}. People have personal preferences including individual interests, habits and behavioral patterns, which often contribute to their next location choices during navigation.
\item {\bf Temporal factors}. People tend to exhibit nonuniform and periodic moving behaviors. For instance, people usually leave home in the morning and return in the evening of weekdays. Therefore, temporal information may be of significant importance and requires proper handling.
\item {\bf Relative importance of factors}. Various factors (personal, sequential and temporal information) may play different roles in affecting human mobility patterns, e.g., sometimes a user's personal preference is more important during navigation, but sometimes the current location dominates the result. Simply ignoring the difference may hinder capturing the real picture of people's mobility patterns.
\end{enumerate}

Current methods mainly adopt Markov or Bayesian models to predict next locations \cite{zhang2016gmove,jia2016location,ye2013s,chen2014nlpmm}, in which the core idea is to compute the conditional probability of each possible next location given current observations (object, location, and time slot) and select the one that has the highest probability as the predicted. The drawbacks of these methods lie in two aspects. 1) \textbf{Improper independence assumption}: methods based on Bayesian models assume that the attributes in the trajectory data are mutually independent, which rarely holds in practice.  2) \textbf{Over-fitting}: methods based on Markov models often suffer from the problem of over-fitting as the number of training instances given the specific observation is limited, e.g., each object appears in only 7 records on average in one day in our VPR data. 

We envision a solution that is able to not only jointly consider these factors (including road network constraints, sequential patterns, personal tendencies and temporal influences), but project objects, locations and time slots into the same low-dimensional latent space, to effectively represent human mobility patterns. This approach has at least three advantages: (1) we could map all the attributes (objects, locations, time) into the same space, without the assumption of independence; (2) for a given attribute, we could use all of the training instances containing this attribute to adjust its position in the space, alleviating the problem of over-fitting; (3) we could compute the correlation between any two points in the space with a distance metric, which would allow us to understand the relationship between different objects (or time), e.g., for a particular object (time slot), which objects (slots) may demonstrate a more similar behavior. In light of recent advances in distributed representation \cite{Le2014Distributed,Wang2016Improving}, we explore the use of embedding methods and aim to accommodate these attributes in a latent space.

In fact, there have already been a few embedding methods that attempt to model check-ins for POI (point of interest) recommendation \cite{feng2015personalized,zhao2017geo}. However, the problem of POI recommendation is notably different from that of next location prediction. For example, the methods modeling check-in POIs usually focus on the activity at the individual POI instead of the visiting order of successive POIs, whereas for predicting next locations, the preceding locations and the order of their visits often play a vital role. Moreover, existing methods for POI recommendation do not consider the restriction of road networks, which renders them unsuitable to be applied to our problem directly.

\textbf{Present work.} We propose a novel Mobility Pattern Embedding (MPE) method to effectively represent human mobility patterns. It considers the joint action of different attributes, and leverages distributed representations to model objects, locations, and time slots jointly. MPE is especially useful for the task of predicting next locations, in that the next location is associated with an embedding vector, and its \textit{conditional vector} can be computed by summing up the embedding vectors of the conditional attributes. Given a next location $l_{i+1}$, its conditional attributes contain the object $o$, the current location $l_{i}$ and the time $t$ that $o$ arrives at $l_i$. The objective is to minimize the Euclidean distances between the embedding vectors of the next locations and their conditional vectors in the latent space. As a result, MPE is general and flexible to model these conditional attributes in a unified way.

Note that, we distinguish between the role of a next location and a current location, and represent the same location using different vectors in the same space depending on which role it takes, to eliminate the effect of \textit{phantom transitions} in trajectories. To illustrate this, consider the following sample scenario, where each location is mapped to a single point in the embedding space, irrespective of its role (current or next location). Given two transitions $l_i \rightarrow l_j$ and $l_j \rightarrow l_k$, if we knew that both $l_i$ and $l_k$ are close to $l_j$ in the space, then it would be of high probability to predict $l_i \rightarrow l_k$. However, it is not likely to observe $l_i \rightarrow l_k$ due to the restriction of road networks, unless there exists a direct road between $l_i$ and $l_k$. Our proposal helps solve this problem by mapping the same location $l_j$ to two different points in the space according to its role.

We exploit the stochastic gradient descent method to estimate the parameters, and conduct thorough experimental studies on two real datasets: the vehicle passage records generated by over 18,000 vehicles from a traffic surveillance system and the publicly available trajectory data of 442 taxis for a complete year. We demonstrate the effectiveness of MPE on the task of next location prediction and visualization of embedding vectors. The experimental results confirm the superiority of our model over alternative methods. The major contributions can be summarized as follows.

\begin{itemize}
\item We propose a novel Mobility Pattern Embedding model by considering the features of traffic trajectory data, i.e., ``phantom transitions'' usually do not exist due to the restriction of road networks. To the best of our knowledge, this is the first work that uses embedding method to model mobility patterns from traffic trajectory data .
\item We consider the sequential, personal and temporal information in a unified way and project objects, time slots, current locations and next locations as points in a low-dimensional latent space to better model human mobility patterns. The availability of such embedding vectors could benefit a wide spectrum of applications such as next location prediction and visualization.
\item We conduct extensive experiments with real VPR data and taxi trajectory data, and compare MPE with baselines on the task of next location prediction. Further, we visualize the embedding vectors of objects and time, and the clear patterns confirm the effectiveness of MPE.
\end{itemize}

The rest of this paper is organized as follows. Section~\ref{relatedwork} reviews the studies on embedding learning, next location prediction and POI recommendation. Section~\ref{pre} introduces the definition of some concepts and the problem solved in this paper. Section~\ref{mpe} presents our mobility pattern embedding model.  The experimental results are discussed in Section~\ref{experiment}. Section~\ref{conclusion} describes the concluding remarks.

\section{Related Work}\label{relatedwork}
Trajectory data mining has been a hot research topic recently with the availability of massive spatial trajectory data \cite{zheng2017popularity,zhu2017effective,jiang2017feature,chen2015mining,zheng2015trajectory}. Zheng \cite{zheng2015trajectory} conducts a systematic survey on this filed, including trajectory data preprocessing, trajectory data management, and a variety of mining tasks. We focus on predicting the next locations of moving objects with embedding methods, so we first discuss the recent progress of embedding learning methods and then review the studies on location prediction and POI recommendation.

\subsection{Embedding Learning}
Embedding objects from high-dimensional vectors into a lower-dimensional space is an important operation in machine learning, and has been successfully exploited in an array of applications including visualization and speech recognition \cite{Hinton2010Stochastic,Graves2014Towards}. Recently, word2vec \cite{mikolov2013distributed} has been proved to be an efficient method for learning high-quality distributed vector representations of words. It models the words' contextual correlations in word sentences, achieving better performance in many natural language processing tasks such as word analogy and machine translation. Meanwhile, similar methods \cite{an2017poi2vec,grover2016node2vec} have been proposed for learning distributed vector representations for nodes in the network and POIs in the physical world. Feng et al. \cite{an2017poi2vec} incorporate the geographical influence into a new latent representation model POI2vec for predicting potential visitors for a given POI. Grover and Leskovec \cite{grover2016node2vec} propose an algorithmic framework named node2vec for learning continuous feature representations for nodes in networks.


\subsection{Location Prediction}

There exist an array of studies that use different methods (e.g., Markov models, frequent patterns) to mine mobility patterns from historical traffic trajectory data to predict the next locations. For example, Monreale et al. \cite{monreale2009wherenext} consider the historical movements of all moving objects to build a T-pattern tree to make future location prediction. Chen et al. \cite{chen2015mining,chen2014nlpmm} propose to mine both individual and collective movement patterns with an integrated variable-order Markov model to predict next locations. Xue et al. \cite{xue2013destination} first decompose historical trajectories into sub-trajectories and connect them into synthesised trajectories, and then use a Markov model to predict the destination of an object.

In addition, there also exist methods that use the neural networks to model human mobility patterns. For instance, De Br{\'e}bisson \textit{et al.} \cite{de2015artificial} introduce an almost fully-automated neural network to predict the destination of a taxi based on both the initial location of the trajectory and its associated meta-data. Liu \textit{et al.} \cite{liu2016predicting} propose a novel method called Spatial Temporal Recurrent Neural Networks (ST-RNN) which models local temporal and spatial contexts in each layer for mining mobility patterns. ST-RNN focuses on storing statistical weights for long-term transitions in a trajectory, whereas we aim at modeling the transitions from current locations to next ones.

Pushing further from the historical trajectories, there are some studies that improve prediction accuracy by taking external information (e.g., semantic information, driving speed) into consideration. For example, Zhou et al. \cite{zhou2013semi} train a local model based on a small set of reference trajectories to predict the future movement of the target object. Zhang et al. \cite{zhang2015nextme} extract the underlying correlation between human mobility patterns and cellular call patterns and make location prediction from temporal and spatial perspectives with it. However, the above methods can only be applied to some specific trajectory data with these external information.

In summary, the existing methods mainly model these attributes (object, conditional location, and time slot) independently, and we cannot understand the relationship between two objects (or time slots) with the discovered patterns. In this study, we choose an angle different from these models, in which we embed all the attributes into the same latent space, and measure the relationship by computing the Euclidean distance of two points.

\subsection{POI Recommendation}
Some recent studies on POI recommendation in location-based social networks are also related to our work, in which any unvisited POIs can be recommended to users. For example, Zhao et al. \cite{Zhao2016stellar} propose a spatial-temporal latent ranking (STELLAR) method based on a ranking-based pairwise interaction tensor factorization framework to make POI recommendation. Yuan et al. \cite{yuan2013and,yuan2015and} present a probabilistic model W4 (short for Who+Where+When+What) to exploit short text messages associated with geographic information, posting time, and user ids to discover user mobility behaviors for POI recommendation. Yao et al. \cite{yao2016poi} propose to incorporate the degree of temporal matching between users and POIs when making personalized POI recommendations. Lian et al. \cite{lian2014geomf} incorporate the spatial clustering phenomenon into weighted  matrix factorization to help improve POI recommendation performance. Yin et al. \cite{yin2015joint} propose a unified generative model to simultaneously model the semantic, temporal and spatial patterns of users' check-in activities for POI recommendation. However, these methods do not have decent performances in predicting next locations, as they fail to consider the just-passed locations, which play pivotal roles in affecting people's decision-making for next locations.


The works \cite{zhou2016general,feng2015personalized, zhao2017geo} that focus on making POI recommendation with embedding vectors are more related to ours. Zhou et al. \cite{zhou2016general} propose a Multi-Context Trajectory Embedding Model (MC-TEM) for POI recommendation, which uses the framework of \textit{word2vec} directly, and takes various useful contextual features, including user-level, trajectory-level, location-level and temporal contexts, into consideration. Feng et al. \cite{feng2015personalized} propose a personalized ranking metric embedding method (PRME), which first embeds each POI into a sequential transition space, and then projects each POI and user into a user preference space. As two components contribute differently in POI recommendation, it uses a linear interpolation to balance them. Zhao et al. \cite{zhao2017geo} assume that the contextual check-in information implies complementary knowledge of POIs, and propose a Geo-Temporal sequential embedding rank (Geo-Teaser) model for POI recommendation. Geo-Teaser first encodes POIs with the framework of \textit{word2vec} by treating each user as a ``document'', check-ins in a day as a ``sentence'', and each POI as a ``word'', and then combines personal and temporal information. However, this method only discriminates weekdays and weekends concerning the temporal factor, failing to consider the subtle variation of different time slots in a day.



In order to highlight our contribution, we summarize the main difference between the proposed model MPE and the aforementioned methods. First, POI recommendation pays little attention to the visiting order of POIs, and ``phantom transitions'' indeed exist in the check-in data. For example, given observations of frequent POI transitions \textit{Home} $\rightarrow$ \textit{Subway} (going from Home to Subway directly) and \textit{Subway} $\rightarrow$ \textit{Market}, it is of high probability to observe \textit{Home} $\rightarrow$ \textit{Market} as well; thus these methods represent the POIs with values in one vector set without distinguishing the current POI and the next one, but this is inappropriate in our problem. Second, Geo-Teaser \cite{zhao2017geo} and MC-TEM \cite{zhou2016general} adopt the framework of \textit{word2vec} directly, and model the correlation between one location and its context (e.g., the previous $K$ and the successive $K$ locations), whereas our MPE directly models the transition (from the current location to the next one), and defines the novel objective function different from \textit{word2vec}. Finally, PRME \cite{feng2015personalized} models the sequential and personal information independently, and does not consider the temporal information. To the best of our knowledge, we are the first to model personal, sequential, and temporal factors simultaneously with the embedding method for traffic trajectory data.

\section{Preliminaries}\label{pre}
We first define some concepts which are required for the subsequent discussion, and introduce the intuitions behind our proposed model. Then we list the notations and their descriptions in Table~\ref{tab:notation}.

\subsection{Concepts}
\begin{myDef}[Record] Each \textbf{\textit{record}} is represented as a triple $r: (o,t,l)$, where $o$ refers to an object ID, $l$ indicates a location ID, and $t$ represents the time-stamp $o$ arrives at $l$.
\end{myDef}

Note that we discretize the time span into equi-sized buckets for simplicity, and represent $t$ with the time slot it belongs to. Further, the size of the time slot is data-independent and can be determined experimentally. Given an object $o$, we sort its records by time, and construct a \textbf{quadruple} $c: (o,t,l_i,l_j)$ for each record $r$, where $l_j$ is the next location the user $o$ will arrive at directly from the location $l_i$ in the time $t$.

\begin{myDef}[Transition] For a quadruple $(o,t,l_i,l_j)$, we define $l_i \rightarrow l_j$ as a \textbf{transition}, meaning an object could arrive at $l_j$ from $l_i$ directly without passing through any other location.
\end{myDef}
Specially, given a transition $l_i \rightarrow l_j$, we define $l_i$ as the current location, and $l_j$ as the next location.

\begin{myDef}[Sequence] For two transitions $l_i \rightarrow l_j$ and $l_j \rightarrow l_k$, we define $l_i \rightarrow l_j \rightarrow l_k$ as a \textbf{sequence}.
\end{myDef}

\begin{myDef}[Candidate Next Location] For a location $l_i$, we define a location $l_j$ as a \textbf{candidate next location} of $l_i$ if the transition $l_i \rightarrow l_j$ exists.
\end{myDef}

\begin{myPro}[Mobility Pattern Embedding] Given the historical quadruples $\cal {C}$, \textbf{Mobility Pattern Embedding} aims at modeling the interactions of objects, time slots, current locations and next locations in a unified way by embedding the four attributes in a latent vector space.
\end{myPro}

\begin{table}
\centering
\caption{Notations and descriptions.}
\renewcommand{\arraystretch}{1.2}
\begin{tabular}{ l| l}
\Xhline{1pt}
Notations & Descriptions \\
\hline
$o$ &  an object (e.g., vehicle, taxi)  \\
$l$  & a location  \\
$t$ &  a time slot\\
$r$  &  a record containing $o$, $l$, $t$\\
$c$ &  a quadruple containing $r$ and its next location\\
$\cal O$ &  the set of objects  \\
${\cal L}^{c}$ &  the set of current locations\\
${\cal L}^{n}$ &  the set of next locations\\
$\cal T$  & the set of time slots  \\
$\cal C$ &  the set of historical quadruples\\
\hline
$D$ &  the embedding's dimensionality\\
$M$ &  the number of negative samples\\
$\mathbf{O}$ &  the distributed representations of objects\\
$\mathbf{L^{c}}$ &  the distributed representations of current locations\\
$\mathbf{L^{n}}$ &  the distributed representations of next locations\\
$\mathbf{T}$ &  the distributed representations of time slots\\
\Xhline{1pt}
\end{tabular}
\label{tab:notation}
\end{table}	

\subsection{Intuitions}
\textbf{Sequential influence}. It has been shown that people's next movements depend on the sequential correlations of successive locations \cite{chen2015mining}, which can be caused by personal navigation habits or the restriction of road networks. In addition, as the movements of vehicles are subject to real road networks, the ``phantom transitions'' rarely occur. That is, given two transitions $l_i \rightarrow l_j$ and $l_j \rightarrow l_k$, $l_i \rightarrow l_k$ does not exist unless there is a direct route from $l_i$ to $l_k$. To validate the transitivity, we analyze two real datasets: VPR data and Taxi data (the detailed descriptions on the data are listed in Section~\ref{experiment}). For the VPR data, we obtain 1,704 transitions and 6,329 sequences, and only 7.46\% of sequences have the characteristic of ``phantom transitions''; for the Taxi data, we obtain 11,645 transitions and 59,949 sequences, and only 10.38\% do. We thus need to avoid the ``phantom transitions'' in the proposed embedding model.

\textbf{Personal tendencies}.
Intuitively, the personal preference which reflects the overall user interests, habits and behavioural patterns may affect the location choices in trajectories. With the same current location, two objects are likely to arrive at different next locations. For example, two persons living in the same apartment might go to lunch in ``McDonald's'' and ``Pizza Hut'' respectively. Therefore, given the current location of an object, the predicted next location should not only be related to the current location, but also capture the object's preference.

\begin{figure}[!t]
\centering
\includegraphics[width=0.5\textwidth]{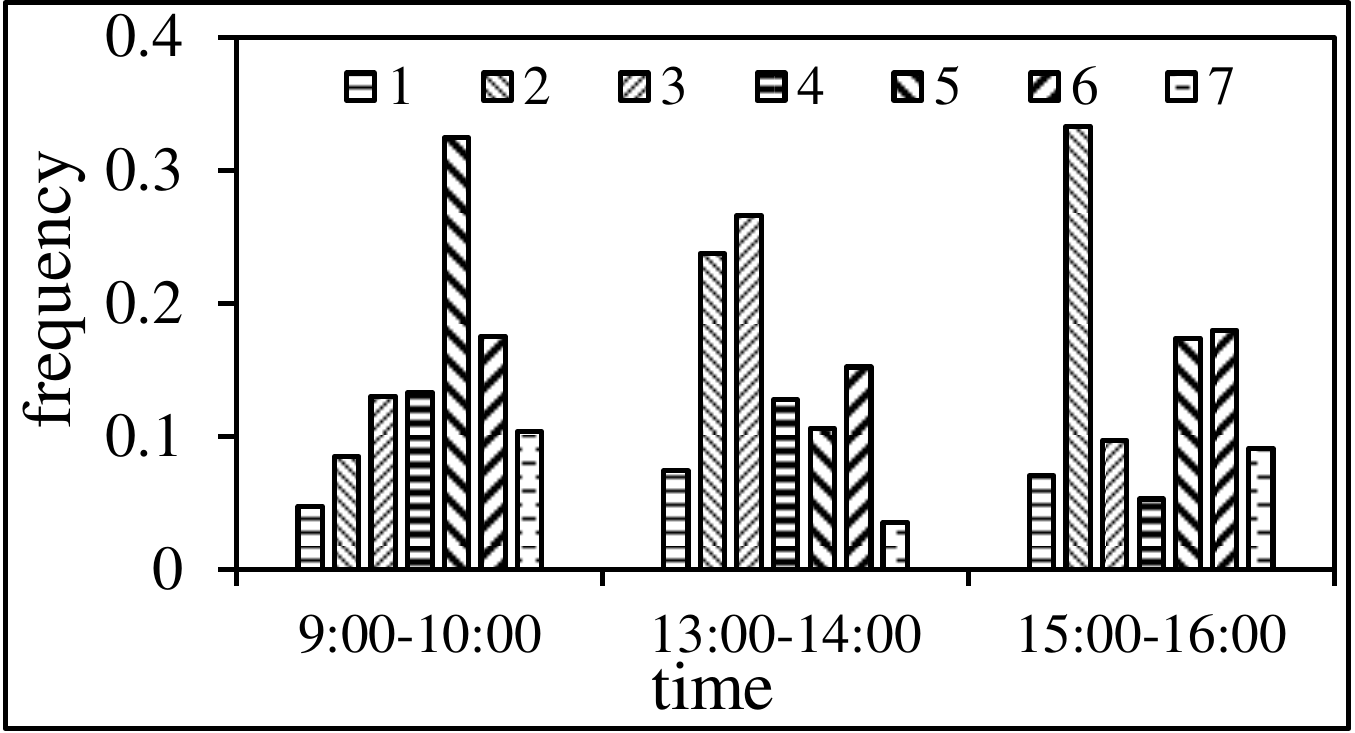}
\caption{An example of time affecting next locations.}
\label{timedistribution}
\end{figure}

\textbf{Temporal influence}.
Different mobility patterns exist in different time slots, e.g., Bob is going to leave home, and he is most likely to go to work at 8 am, and have lunch at 11:30 am. To illustrate how time could affect people's decisions on next locations, we sample a location in the VPR data. Note that it has seven candidate next locations, and the distributions over those locations do differ from one period to another, as shown in Fig.~\ref{timedistribution}. For instance, vehicles are most likely to arrive at the fifth location during the period from 9:00 to 10:00, whereas the most probable next location is the second for the period from 15:00 to 16:00. We thus should differentiate hours of a day to reflect the temporal influence.

\section{Mobility Pattern Embedding Model}\label{mpe}
In this section, we first present a Mobility Pattern Embedding (MPE) model, and then introduce the parameter learning algorithm and its complexity analysis.

\subsection{Model Description}
We propose a novel Mobility Pattern Embedding (MPE) method to model the combined action of sequential, personal and temporal influences on people's mobility patterns. MPE embeds objects, time slots, current locations and next locations together as points in a low-dimensional latent space. Specifically, $\mathbf{O} \in \mathbb{R}^{|{\cal O}| \times D}$ is the object embedding matrix, $\mathbf{T} \in \mathbb{R}^{|{\cal T}| \times D}$ is the time embedding matrix, $\mathbf{L^c} \in \mathbb{R}^{|{\cal L}^{c}| \times D}$ is the current location embedding matrix, $\mathbf{L^n} \in \mathbb{R}^{|{\cal L}^{n}| \times D}$ is the next location embedding matrix, where $D$ is the embedding's dimensionality, $\cal O$, $\cal T$, ${\cal L}^{c}$ and ${\cal L}^{n}$ are the sets of objects, time slots, current locations and next locations, respectively.

For a quadruple $c: (o,t,l_i,l_j)$ where $r = (o,t,l_i)$, we define the \textit{conditional vector} $\vec{V^{c}}$ of the next location $l_j$ as the sum of three vectors $\mathbf{O}_o$, $\mathbf{T}_t$ and $\mathbf{L^c}_{l_i}$, i.e., $\vec{V^{c}} = \mathbf{O}_o + \mathbf{T}_t + \mathbf{L^{c}}_{l_i}$, where $\mathbf{O}_o$ is the embedding vector of the object $o$, $\mathbf{T}_t$ is the embedding vector of the time slot $t$, and $\mathbf{L^{c}}_{l_i}$ is the embedding vector of the current location $l_i$. Here we assume that the Euclidean distance between the vector $\mathbf{L^n}_{l_j}$ of the next location $l_j$ and its conditional vector $\vec{V^{c}}$ reflects the transition probability of $r$ to $l_j$, and its value can be estimated as:
\begin{equation}
\label{firstprob}
\hat{P}(l_j|r)=\dfrac{\exp(- \| \mathbf{L^n}_{l_j}-\vec{V^{c}}\| ^{2})}{Z(\vec{V^{c}})},
\end{equation}
where $Z(\vec{V^{c}})$ is the normalization term.

In order to compute $\hat{P}(l_j|r)$, we adopt the method of negative sampling \cite{mikolov2013distributed,Levy2014Neural}, and maximize $\hat{P}(l_j|r)$ for the observed next location $l_j$ while minimizing $\hat{P}(l|r)$ for the randomly sampled unobserved (negative) next location $l$. The objective of MPE for a single tuple $c: (r,l_j)$ therefore becomes:
\begin{equation}
\max \left( \hat{P}(l_j|r) -  \sum_{l_{m} \in NEG(r)} \hat{P}(l_m|r) \right),
\end{equation}
where $NEG(r)$ is the set of unobserved next locations for record $r$, $l_{m} \in NEG(r)$ is the sampled next location, and the number of ``negative'' samples is $M$.

Actually, we are concerned only with their ranking, instead of estimating the probability of each possible next location. For example, if tuple $(r,l_{j})$ is observed and $(r,l_{m})$ is unobserved, we expect that the value of probability $\hat{P}(l_{j}|r)$ should be higher than $\hat{P}(l_{m}|r)$.  Accordingly, we can simplify the computation by keeping only the Euclidean distance instead of applying the exponential function:

\begin{equation}
\begin{aligned}
&\hat{P}(l_{j}|r) > \hat{P}(l_{m}|r) \\
&\Rightarrow \exp\left(-\|\mathbf{L^n}_{l_j}-\vec{V^{c}}\|^{2}\right) > \exp\left(-\|\mathbf{L^n}_{l_m}-\vec{V^{c}}\|^{2}\right)\\
&\Rightarrow \|\mathbf{L^n}_{l_m}-\vec{V^{c}}\|^{2} -\|\mathbf{L^n}_{l_j}-\vec{V^{c}}\|^{2} > 0.
\end{aligned}
\end{equation}

Therefore, given a quadruple $c: (r,l_j)$, we randomly sample $M$ unobserved next location $l_{m} \in NEG(r)$, and expect that $\hat{P}(l_{j}|r)$ should
be higher than $\hat{P}(l_{m}|r)$. We could redefine the objective with the maximum likelihood function:
\begin{equation}
\begin{aligned}
 &\max \prod_{l_{m} \in NEG(r)} P\left(\hat{P}(l_{j}|r) > \hat{P}(l_{m}|r)\right)\\
\Rightarrow &\max \prod_{l_{m} \in NEG(r)} P \left( \|\mathbf{L^n}_{l_m}-\vec{V^{c}}\|^{2} -\|\mathbf{L^n}_{l_j}-\vec{V^{c}}\|^{2}>0 \right).
\end{aligned}
\end{equation}

A sigmoid function has a domain of all real numbers with return value monotonically increasing from 0 to 1 and is differentiable having a non-negative first derivative. Using the sigmoid function $\sigma(z) = (1+\exp(-z))^{-1}$ , the objective can be further written as
\begin{equation}
\max \prod_{l_{m} \in NEG(r)} \sigma \left(\|\mathbf{L^n}_{l_m}-\vec{V^{c}}\|^{2} -\|\mathbf{L^n}_{l_j}-\vec{V^{c}}\|^{2}\right).
\end{equation}

Taking the log function on the objective, we have
\begin{equation}
\begin{aligned}
&\max \log \prod_{l_{m} \in NEG(r)} \sigma \left(\|\mathbf{L^n}_{l_m}-\vec{V^{c}}\|^{2} -\|\mathbf{L^n}_{l_j}-\vec{V^{c}}\|^{2}\right)\\
\Rightarrow &\max \sum_{l_{m} \in NEG(r)} \log \sigma \left(\|\mathbf{L^n}_{l_m}-\vec{V^{c}}\|^{2} -\|\mathbf{L^n}_{l_j}-\vec{V^{c}}\|^{2}\right).
\end{aligned}
\end{equation}

Finally, the objective $\ell$ of MPE for all the tuples can be described as:
\begin{equation}
\begin{aligned}
\max & \sum_{c=(r,l_j) \in \cal C}  \sum_{l_{m} \in NEG(r)} \log \sigma \left(\|\mathbf{L^n}_{l_m}-\vec{V^{c}}\|^{2} -\|\mathbf{L^n}_{l_j}-\vec{V^{c}}\|^{2}\right),\\
& \vec{V^{c}} = \mathbf{O}_o + \mathbf{T}_t + \mathbf{L^{c}}_{l_i}.
\end{aligned}
\end{equation}

The benefits of MPE are two-fold. (1) MPE distinguishes between the role of a next location and a current location, and represent the same location using two different vectors ($\mathbf{L^{c}}$ and $\mathbf{L^{n}}$) depending on which role it takes; therefore the problem that arises from ``phantom transitions" can be effectively avoided. That is, given $l_i \rightarrow l_j$ and $l_j \rightarrow l_k$ indicating two frequent transitions, both $l_i$ and $l_k$ can be close to the related $l_j$ in the latent space, but with MPE the distance between $l_i$ and $l_k$ can still be vast in the latent space, as the two locations $l_j$ are represented with different vectors depending on their role. (2) We assume that people's movements reflect the combined action of three factors including object, time, and the current location. Instead of manually setting a fixed weight to each factor, MPE automatically estimates all parameters at the same time.

\subsection{Parameter Learning}
The variables in MPE are $\mathbf{\Theta} = (\mathbf{O}, \mathbf{T}, \mathbf{L^{c}}, \mathbf{L^{n}})$, which are parameterized by the fixed $|{\cal O}| \times D$, $|{\cal T}| \times D$, $|{\cal L}^c| \times D$ and $|{\cal L}^n| \times D$ matrix respectively. We learn the MPE using maximum a posterior (MAP):
\begin{equation}
\begin{aligned}
\mathbf{\Theta} &= \arg\max \sum_{c=(r,l_j) \in \cal C}  \sum_{l_{m} \in NEG(r)} \log \sigma \left(\|\mathbf{L^n}_{l_m}-\vec{V^{c}}\|^{2} -\|\mathbf{L^n}_{l_j}-\vec{V^{c}}\|^{2}\right) \\
 &- \lambda\|\mathbf{\Theta}\|^{2},
 \end{aligned}
\end{equation}
where $\lambda\|\mathbf{\Theta}\|^{2}$ is the regularization term.

Here we choose to use stochastic gradient descent to estimate the parameters. Based on the historical data, we obtain a set of quadruples $(o,t,l_{i},l_{j})$, and then randomly sample $M$ unobserved next locations $l_{m}$ for each quadruple. Given a training instance  $(o,t,l_{i},l_{j},l_{m})$, the update
procedure is as follows.
\begin{equation}
\label{eq:update}
\begin{aligned}
\mathbf{O}_o &\leftarrow \mathbf{O}_o + 2 \gamma \left( \left(1-\sigma(z)\right) (\mathbf{L^{n}}_{l_j}-\mathbf{L^{n}}_{l_m}) - \lambda \mathbf{O}_o \right)   \\
\mathbf{T}_t &\leftarrow \mathbf{T}_t + 2 \gamma \left( \left(1-\sigma(z)\right) (\mathbf{L^{n}}_{l_j}-\mathbf{L^{n}}_{l_m}) - \lambda \mathbf{T}_t \right)   \\
\mathbf{L^{c}}_{l_i} &\leftarrow \mathbf{L^{c}}_{l_i} + 2 \gamma \left( \left(1-\sigma(z)\right) (\mathbf{L^{n}}_{l_j}-\mathbf{L^{n}}_{l_m}) - \lambda \mathbf{L^{c}}_{l_i} \right)   \\
\mathbf{L^{n}}_{l_j} &\leftarrow \mathbf{L^{n}}_{l_j} + 2 \gamma \left( \left(1-\sigma(z)\right) (\mathbf{O}_o+ \mathbf{T}_t+ \mathbf{L^{c}}_{l_i} - \mathbf{L^{n}}_{l_j}) - \lambda \mathbf{L^{n}}_{l_j} \right)   \\
\mathbf{L^{n}}_{l_m} &\leftarrow \mathbf{L^{n}}_{l_m} + 2 \gamma \left( \left(1-\sigma(z)\right) (\mathbf{L^{n}}_{l_m}-\mathbf{O}_o - \mathbf{T}_t - \mathbf{L^{c}}_{l_i}) - \lambda \mathbf{L^{n}}_{l_m} \right),
\end{aligned}
\end{equation}
where $z = \|\mathbf{L^{n}}_{l_m}-(\mathbf{O}_o + \mathbf{T}_t + \mathbf{L^{c}}_{l_i})\|^{2} -\|\mathbf{L^{n}}_{l_j}-(\mathbf{O}_o + \mathbf{T}_t + \mathbf{L^{c}}_{l_i})\|^{2}$ and $\gamma$ is the learning rate.

The learning algorithm of MPE is depicted in Algorithm~1. We first initialize the parameters with a Gaussian distribution (Line 1). For each quadruple $(o,t,l_i,l_j)$, we then randomly sample $M$ ``negative'' next locations, and update these parameters based on Equation~(\ref{eq:update}) (Line 3 - 10). We iterate this procedure until the value of $\ell$ remains stable, and finally obtain the approximated optimal parameters. The time complexity of MPE is $O(MDI|\mathcal{C}|)$, where $I$ is the number of iterations, $|\cal{C}|$ is the number of training quadruples, $M$ is the number of ``negative'' samples, and $D$ is the embedding's dimensionality.

\begin{algorithm}[h]
\label{alg:pem}
\caption{Learning Algorithm for MPE}
\begin{algorithmic}[1]
\REQUIRE training quadruples $\cal C$, learning rate $\gamma$, regularization parameter $\lambda$, the number of negative samples $M$, the embedding's dimensionality $D$;
\ENSURE model parameters $\mathbf{O}, \mathbf{T}, \mathbf{L^{c}}, \mathbf{L^{n}}$;
\STATE Initialize the parameters with a Gaussian distribution $N(0,0.01)$;
\REPEAT
\FOR {$c:(o,t,l_i,l_j) \in \cal C$}
\STATE count=0;
\WHILE {$count<M$}
\STATE randomly sample an unobserved next location $l_m \in NEG(r)$;
\STATE update $\mathbf{O}, \mathbf{T} , \mathbf{L^{c}} , \mathbf{L^{n}}$ according to Equation~(\ref{eq:update});
\STATE count++;
\ENDWHILE
\ENDFOR
\UNTIL {stopping criteria is met;}
\STATE \textbf{return} $\mathbf{O}, \mathbf{T}, \mathbf{L^{c}}, \mathbf{L^{n}}$;
\end{algorithmic}
\end{algorithm}

\section{Performance Evaluation}\label{experiment}
We first present experiments using two real datasets to evaluate our proposal with the application of next location prediction, and then visualize the embedding vectors to further confirm the effectiveness of MPE.

\subsection{Datasets and Settings}
In the experiments, we use two datasets: the VPR data and the publicly available taxi trajectory data \footnote{The detailed information about the data can be found here https://www.kaggle.com/c/pkdd-15-predict-taxi-service-trajectory-i/data}.

\begin{table}
\centering
\caption{Data statistics.}
\renewcommand{\arraystretch}{1.2}
\begin{tabular}{ l| l | l}
\Xhline{1pt}
 & VPR data & Taxi data\\
\hline
$\sharp$objects &  34,734 &  442 \\
$\sharp$locations  & 681  &  3,719\\
$\sharp$transitions &  1,704 & 11,645\\
$\sharp$records &  7,205,617  &  32,281,729\\
avg. $\sharp$records of each object &  207.5 & 73035.6\\
density (loc/sq.km) &  0.32  &  9.56 \\
\Xhline{1pt}
\end{tabular}
\label{tab:datastatistic}
\end{table}

\textbf{VPR data}: We collect four weeks (04/01/2016 - 31/01/2016) of VPRs over the traffic surveillance system in a major metropolitan city with an area of 2,119 sq.km. In our dataset, the accuracy of plate number recognition by OCR could reach 97\% in ideal weather/lighting conditions, and we only keep those captured during the daytime (from 7:00 to 17:00) to ensure the data quality. The random recognition errors may result in incomplete/erroneous sequences, and each of such sequences has a low occurrence frequency. We remove such sequences by setting an occurrence threshold of 30 (i.e., each sequence must occur at least 30 times to be included in the dataset), and finally obtain 7,170,883 quadruples in total. Note that, as a side effect of removing the above mentioned errors, we have also removed all instances of rare transitions, as it is difficult to know whether these rare transitions are errors or not.

\textbf{Taxi data}: The taxi data is composed of all the complete trajectories of 442 taxis running in the city of Porto (Portugal) of 389 sq.km for a complete year (from 01/07/2013 to 30/06/2014). We discretize the region of interest into a grid with equal-sized cells, and assign a cell index for each GPS location. After the preprocessing, it generates 32,281,287 quadruples.

\begin{figure}[!t]
\centering
\subfigure[Distribution of the number of records per object.]{
\includegraphics[height=0.13\textheight]{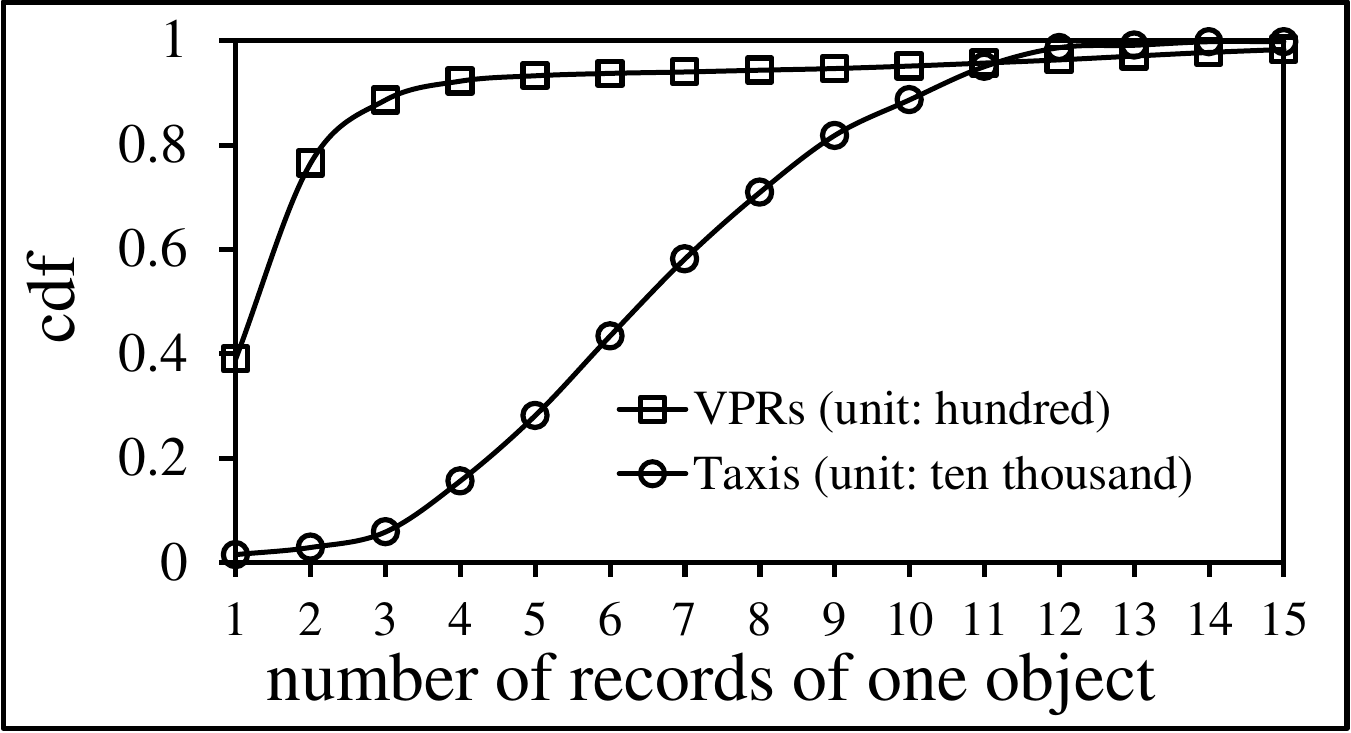}}
\hspace{0.05\textwidth}
\subfigure[Distribution of the number of candidate next locations.]{
\includegraphics[height=0.13\textheight]{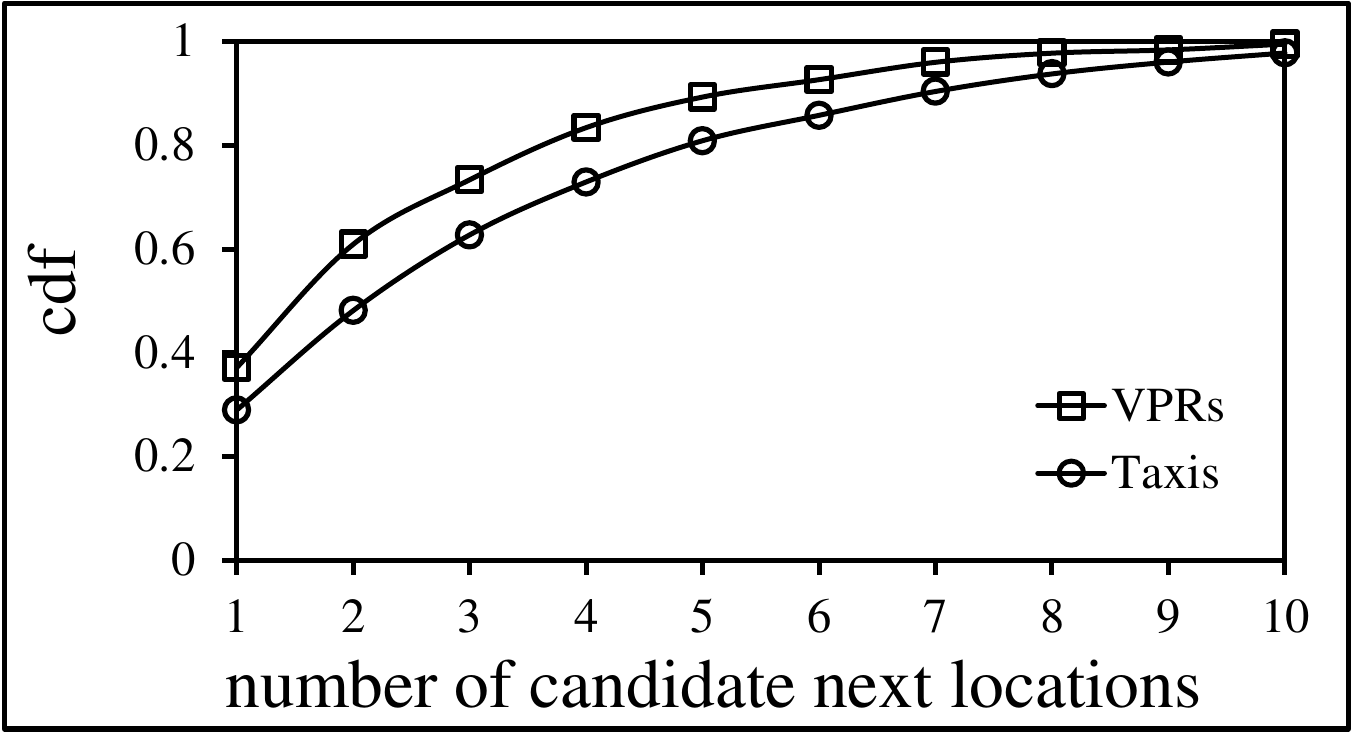}}
\caption{Characteristics of VPR and Taxi data.}
\label{fig:data}
\end{figure}

The statistical properties on both data are shown in Table~\ref{tab:datastatistic}, where $\sharp$objects represents the number of objects and avg. $\sharp$records is the average number of records. We then conduct data analysis to better understand the traffic trajectory data. The cumulative distribution functions (cdf) of the number of records per object and the number of candidate next locations are shown in Fig.~\ref{fig:data}. It can be seen from the figure: (1) for the Taxi data, about 97\% of objects have more than 20,000 records, and about 87\% objects have less than 300 records for the VPR data; (2) it has more candidate next locations on average in the Taxi data than in the VPR data.

For both datasets, we randomly split the quadruples into three collections in proportion of 8:1:1 as the training set, validation set, and test set, and perform 10 runs (with the same data split) to report the average of the results. All the experiments are done on a 3.4GHz Intel Core i7 PC with 16GB main memory. The default values for the number of iterations $I$, the regularization parameter $\lambda$, the embedding's dimensionality $D$, and the number of negative samples $M$ are 10, $10^{-3}$, 100 and 1, and the learning rate $\gamma$ is set at $10^{-3}$. We will evaluate the effect of these parameters in the experiments.

\subsection{Model Convergence and Running Time}
\begin{figure*}[!t]
\centering
\subfigure[VPR data]{
\includegraphics[height=0.13\textheight]{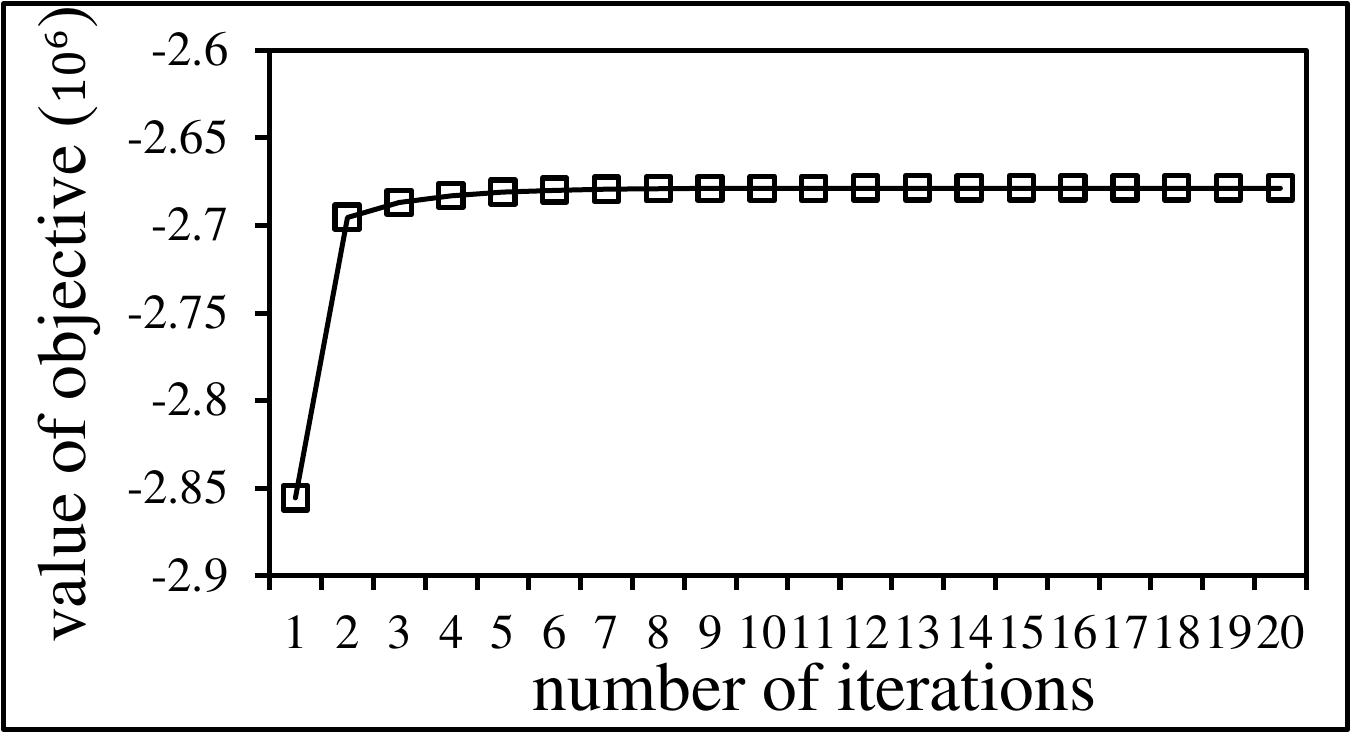}}
\hspace{0.05\textwidth}
\subfigure[Taxi data]{
\includegraphics[height=0.13\textheight]{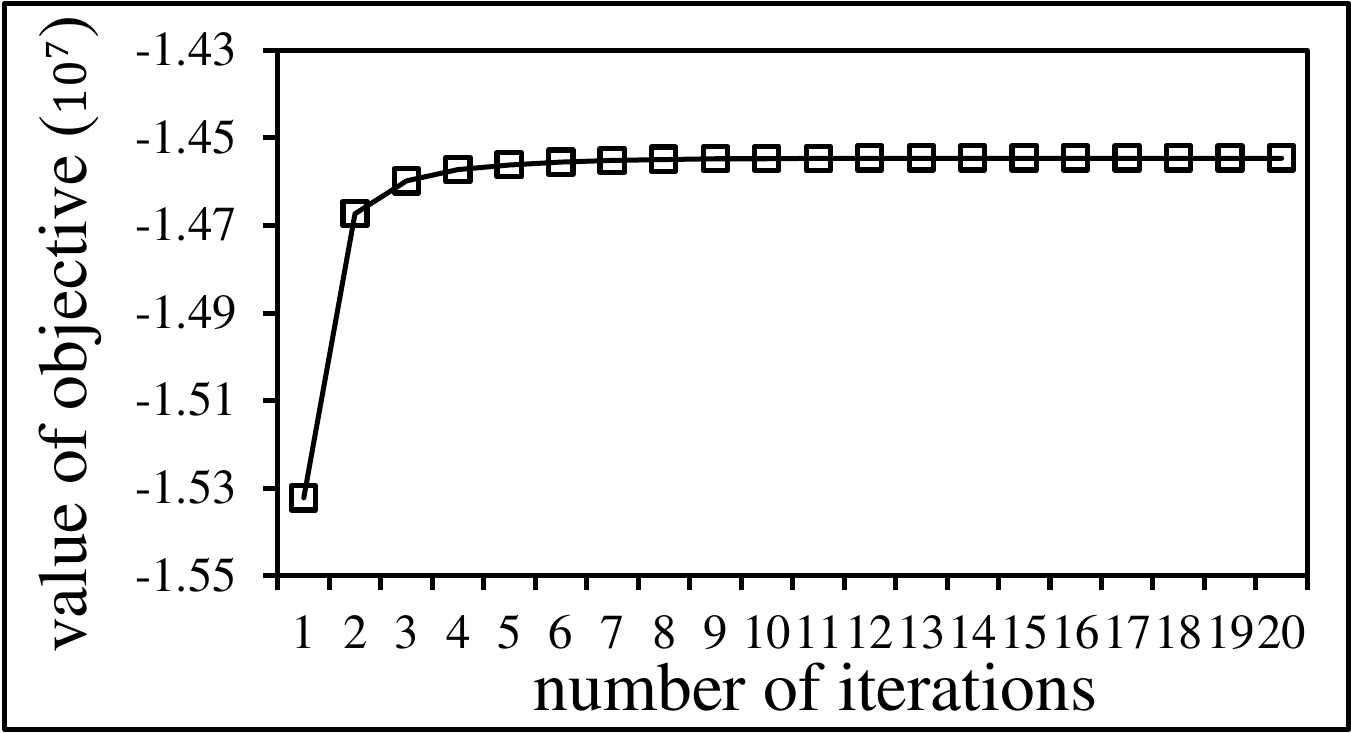}}
\caption{Model Convergence.}
\label{fig:obj}
\end{figure*}

We first validate whether our model's objective achieves a stationary point when iteratively performing these updates. The values of objective function $\ell$ with varying the number of iterations from 1 to 20 on both datasets are shown in Fig.~\ref{fig:obj}. Clearly, with the increase of the number of iterations, the values of $\ell$ increase gradually, and remain stable after about 10 iterations. Hence we set the number of iterations at 10 in the following experiments.

At each iteration, our model needs to update all the parameters, including object embedding matrix $\mathbf{O}$, current location embedding matrix $\mathbf{L^c}$, next location embedding matrix $\mathbf{L^n}$, and time slot embedding matrix $\mathbf{T}$. The size of these matrices and the number of quadruples determine the runtime of each iteration. Table~\ref{table:runningtime} shows the runtime of one iteration for both datasets with different embedding's dimensionality $D$. On one hand, as the Taxi dataset has more quadruples, its runtime is larger than that with the VPR dataset for the same $D$; on the other hand, the runtime increases gradually when we rise $D$. Note that, we could train MPE offline in advance, and use the learned embeddings in the real-time applications.

\begin{table}[!t]
\small
\caption{Runtime of One Iteration (unit: second)}
\renewcommand{\arraystretch}{1.2}
\centering
\begin{tabular}{c  c  c  }
\Xhline{1pt}
 number of vector dimensionality & VPR data & Taxi data\\
\hline
10 & 3.7 & 15.6  \\
50 & 5.8 & 24.4  \\
100 & 8.8 & 41.2  \\
200 & 15.0 & 69.7  \\
300 & 21.8 & 99.6  \\
\Xhline{1pt}
\end{tabular}
\label{table:runningtime}
\end{table}

\subsection{Evaluation on Next Location Prediction}
Given a record $r$ with object, current location and time, the task of \textit{next location prediction} is to predict the most likely successive location. With the proposed MPE, we first build the conditional vector, and then compute $\hat{P}(l_j|r)$ based on Equation~(\ref{firstprob}) for each possible next location $l_j$. Finally, we choose the location with the maximum probability as the predicted next location.

\subsubsection{Baselines}
We compare with the following state-of-the-art methods for predicting next locations to evaluate the performance.
\begin{itemize}
\item \textbf{MM:} the Markov model \cite{chen2014nlpmm}, which mines the mobility patterns for each object with its trajectories to predict next locations.
\item \textbf{Bayes:} it computes the transition probability from $r$ to $l_j$ using Bayes' rules under the assumption that the elements (object, location, and time) of $r$ are independent:
\begin{equation}
\begin{aligned}
\small
P(l_j|r) &=P(l_j|o,l_i,t)\\
           &\propto P(o,l_i,t|l_j)P(l_j)\\
           &\propto P(o|l_j)P(l_i|l_j)P(t|l_j)P(l_j).
\end{aligned}
\end{equation}
\item \textbf{PRME:} the personalized ranking metric embedding method \cite{feng2015personalized}, which considers both sequential information and user preference in training embedding vectors.
\item \textbf{Geo-Teaser:} the geo-temporal sequential embedding rank model \cite{zhao2017geo}, which incorporates personal and temporal information into word2vec framework.
\item \textbf{MC-TEM:} the multi-context trajectory embedding model \cite{zhou2016general}, which takes user-level, trajectory-level, location-level and temporal contexts into consideration.
\item \textbf{MPE-plain:} the simplified MPE model, which just considers the sequential information.
\item \textbf{MPE-object:} the simplified MPE model, which considers the sequential and personal information.
\item \textbf{MPE-time:} the simplified MPE model, which considers the sequential and temporal information.
\end{itemize}

Among the competing methods, MM and Bayes are the popular mobility pattern mining models for next location prediction, which mainly compute the conditional probability; PRME, Geo-Teaser and MC-TEM are relatively advanced embedding models for POI recommendation by mining check-in data; MPE-plain, MPE-object and MPE-time are the simplified versions of our proposed MPE, which only consider part of the factors (sequential, personal and temporal information) that affect people's next locations. 

\subsubsection{Evaluation metrics}
To evaluate the prediction performance, we exploit two well known metrics, namely, \textit{accuracy} and \textit{average precision}. \textit{Accuracy} is defined as $\sum P(l)/|{\cal C}_t|$, where $|{\cal C}_t|$ is the number of quadruples in the test set, and $P(l)$ is 1 if $l$ is the true successive location and 0 otherwise. \textit{Average precision} is defined as $\sum (P(l_w)/w)/|{\cal C}_t|$, where $w$ denotes the position in the predicted list, and $P(l_w)$ takes the value of 1 if $l_w$ is the actual next location and 0 otherwise.

\subsubsection{Experimental results}
\begin{table*}
\caption{Results of methods on VPR data.}
\label{tab:tabel3}
\renewcommand{\arraystretch}{1.2}
\small
\centering
\begin{tabular}{l |c c c |c c c }
\Xhline{1pt}
\multirow{2}*{method} & \multicolumn{3}{c|}{accuracy} & \multicolumn{3}{c}{average precision}   \\
\multirow{2}*{} & top-1 & top-2 & top-3 & top-1 & top-2 & top-3  \\
\hline
MM & 0.543 & 0.635 & 0.660 & 0.543 & 0.589 & 0.597    \\
					
Bayes & 0.634 & 0.814 & 0.894 & 0.634 & 0.724 & 0.751   \\

\hline

PRME & 0.509 & 0.733 & 0.824 & 0.509 & 0.621 & 0.651  \\

Geo-Teaser & 0.539 & 0.734 & 0.822 & 0.539 & 0.636 & 0.666   \\

MC-TEM & 0.534 & 0.728 & 0.819 & 0.534 & 0.631 & 0.659   \\

\hline

MPE-plain & 0.593 & 0.794 & 0.885 & 0.593 & 0.694 & 0.724    \\

MPE-object & 0.633 & 0.828 & 0.903 & 0.633 & 0.732 & 0.754    \\

MPE-time & 0.618 & 0.816 & 0.893 & 0.618 & 0.714 & 0.739   \\

\textbf{MPE} & \textbf{0.645} & \textbf{0.837} & \textbf{0.914} & \textbf{0.645} & \textbf{0.741} & \textbf{0.766}   \\

\Xhline{1pt}
\end{tabular}
\end{table*}

We compare MPE with the baselines using the optimal parameters on VPR data and Taxi data and show the prediction performance in Table~\ref{tab:tabel3} and Table~\ref{tab:tabel4}. The best accuracies and average precisions are highlighted in boldface.
\begin{enumerate}
\item All the methods perform better on the VPR data than on the Taxi data, as the routes taken by taxis are more diverse/random and they may arrive at more candidate next locations (see Fig.~\ref{fig:data}(b)).
\item MM performs the worst on the VPR data due to the limited number of records of each object (see Fig.~\ref{fig:data}(a)), but gets decent top-1 accuracy and average precision on the Taxi data, as it has sufficient records to capture individual mobility patterns. Bayes takes objects, current locations and time slots into consideration, and performs much better than MM. Our proposed MPE models the same factors as Bayes, and it performs better, as MPE considers the combined action of these factors instead of treating them independently.
\item PRME, Geo-Teaser and MC-TEM represent both current locations and next locations with the same vector set based on the assumption of ``phantom transitions'', which is not applicable to the traffic trajectory data, limiting their prediction performance. Our proposed MPE outperforms them significantly, for instance, compared with Geo-Teaser, which has the best performance among the three methods, the top-3 accuracy and average precision improve by 11.2\% and 15.0\% respectively on the VPR data, and by 18.0\% and 17.8\% on the Taxi data. The reasons lie in two-fold: on one hand, MPE prevents the ``phantom transitions'' by distinguishing current locations and next ones; on the other hand, MPE is capable of learning the human mobility patterns by modeling the interactions of personal, sequential and temporal influences in a unified way.
\item MPE-plain, MPE-object and MPE-time just consider part of the factors in modeling human mobility patterns, and they perform worse than MPE. Compared with MPE-plain, MPE-object and MPE-time model the object and time information respectively, and they obtain decent performances. Further, MPE-object has higher prediction accuracies, indicating that personal information plays a more important role in affecting people's mobility patterns.
\end{enumerate}

\begin{table*}
\caption{Results of methods on Taxi data.}
\label{tab:tabel4}
\renewcommand{\arraystretch}{1.2}
\small
\centering
\begin{tabular}{l |c c c |c c c }
\Xhline{1pt}
\multirow{2}*{} & \multicolumn{3}{c|}{accuracy} & \multicolumn{3}{c}{average precision}  \\
\multirow{2}*{} &  top-1 & top-2 & top-3 & top-1 & top-2 & top-3 \\
\hline
MM &  0.392 & 0.610 & 0.739 & 0.392 & 0.501 & 0.544  \\
					
Bayes   & 0.393 & 0.614 & 0.750 & 0.393 & 0.504 & 0.548  \\

\hline

PRME &  0.321 & 0.506 & 0.641 & 0.321 & 0.413 & 0.459 \\

Geo-Teaser &  0.336 & 0.525 & 0.651 & 0.336 & 0.431 & 0.472  \\

MC-TEM &  0.332 & 0.518 & 0.644 & 0.332 & 0.423 & 0.466  \\

\hline

MPE-plain & 0.379 & 0.603 & 0.732 & 0.379 & 0.483 & 0.524  \\

MPE-object & 0.391 & 0.619 & 0.756 & 0.391 & 0.502 & 0.547  \\

MPE-time & 0.386 & 0.611 & 0.745 & 0.386 & 0.497 & 0.538  \\

\textbf{MPE} &  \textbf{0.397} & \textbf{0.633} & \textbf{0.768} & \textbf{0.397} & \textbf{0.507} & \textbf{0.556}  \\

\Xhline{1pt}
\end{tabular}
\end{table*}

\subsubsection{Parameter setting and tuning}\label{para}

\begin{figure*}[!t]
\centering
\subfigure[VPR data]{
\includegraphics[height=0.13\textheight]{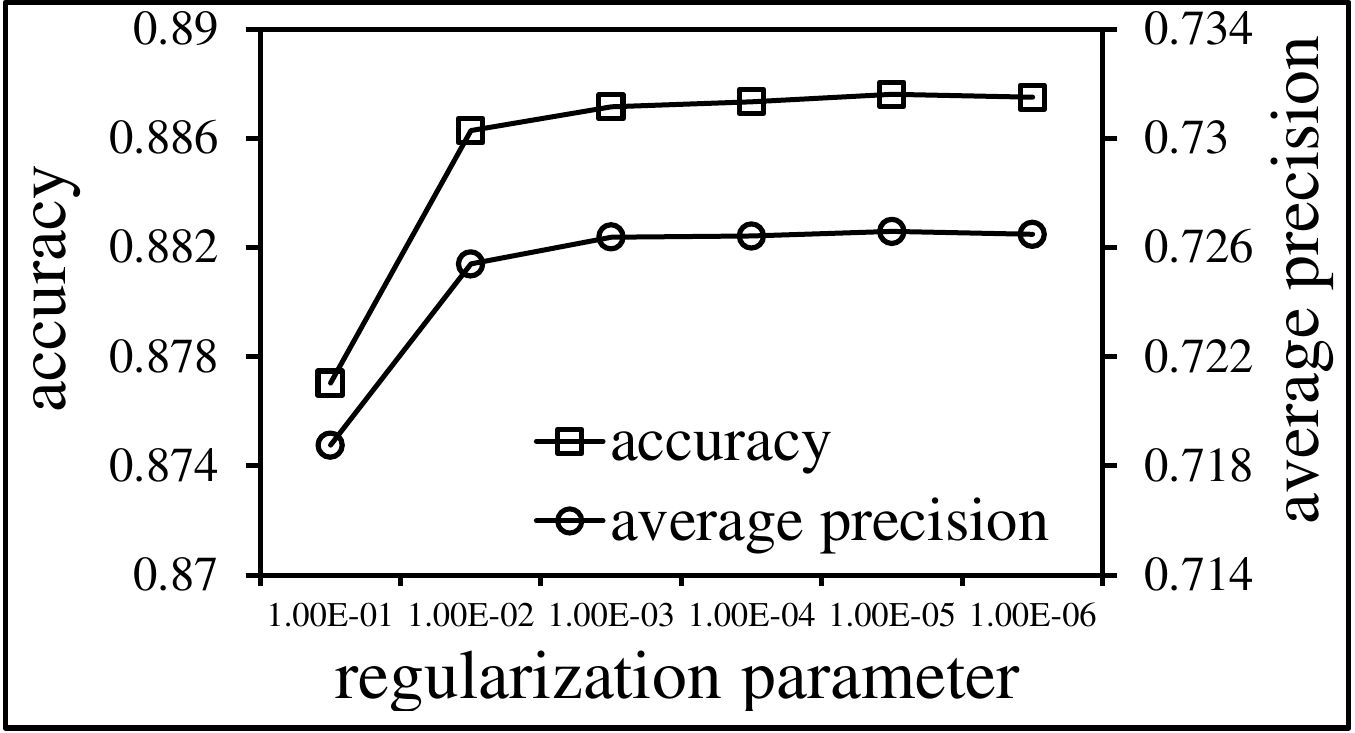}}
\hspace{0.05\textwidth}
\subfigure[Taxi data]{
\includegraphics[height=0.13\textheight]{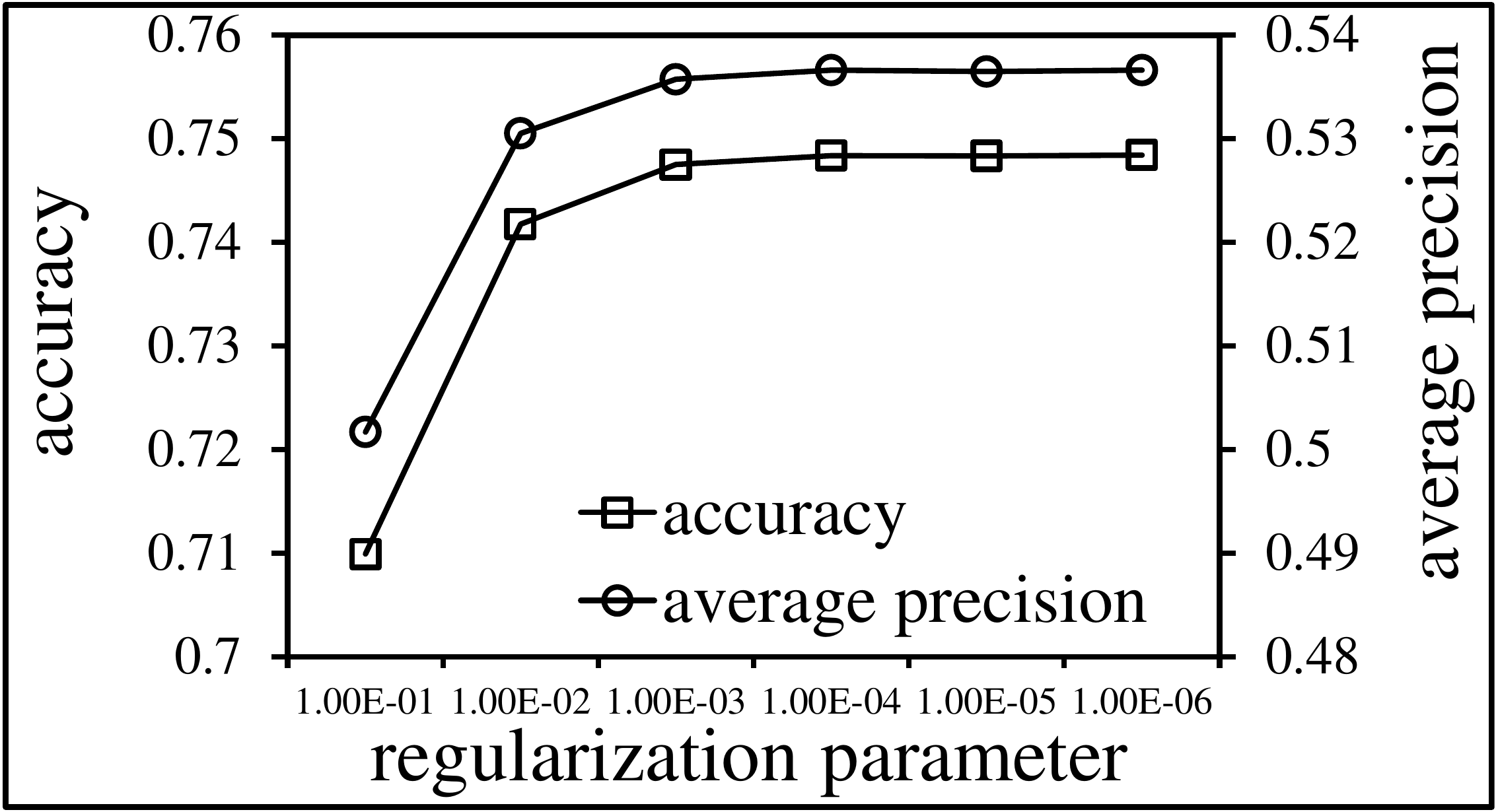}}
\caption{Effect of the regularization parameter $\lambda$.}
\label{fig:reg}
\end{figure*}

\begin{figure*}[!t]
\centering
\subfigure[VPR data]{
\includegraphics[height=0.13\textheight]{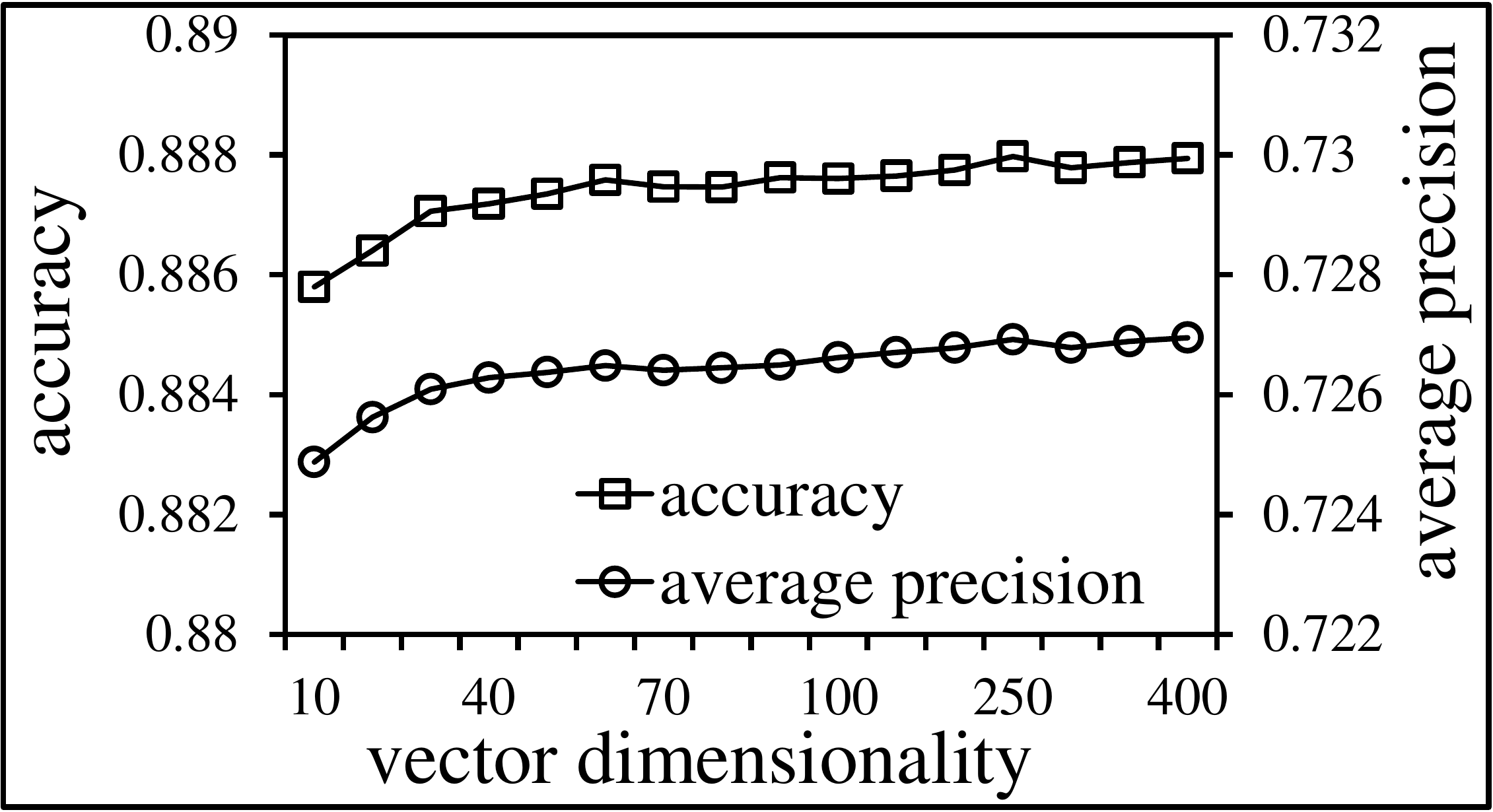}}
\hspace{0.05\textwidth}
\subfigure[Taxi data]{
\includegraphics[height=0.13\textheight]{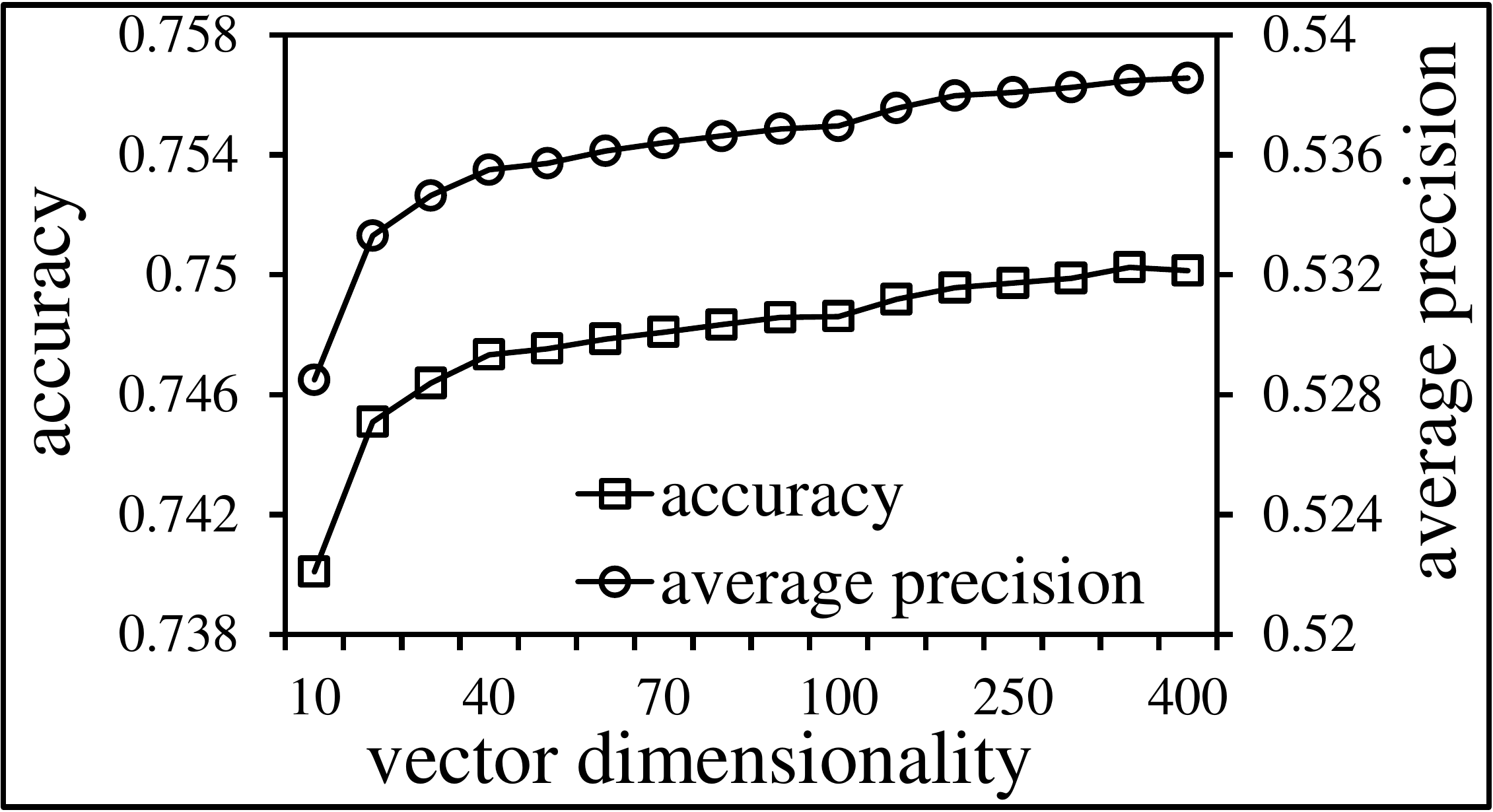}}
\caption{Effect of the embedding's dimensionality $D$.}
\label{fig:dim}
\end{figure*}

\begin{figure*}[!t]
\centering
\subfigure[VPR data]{
\includegraphics[height=0.13\textheight]{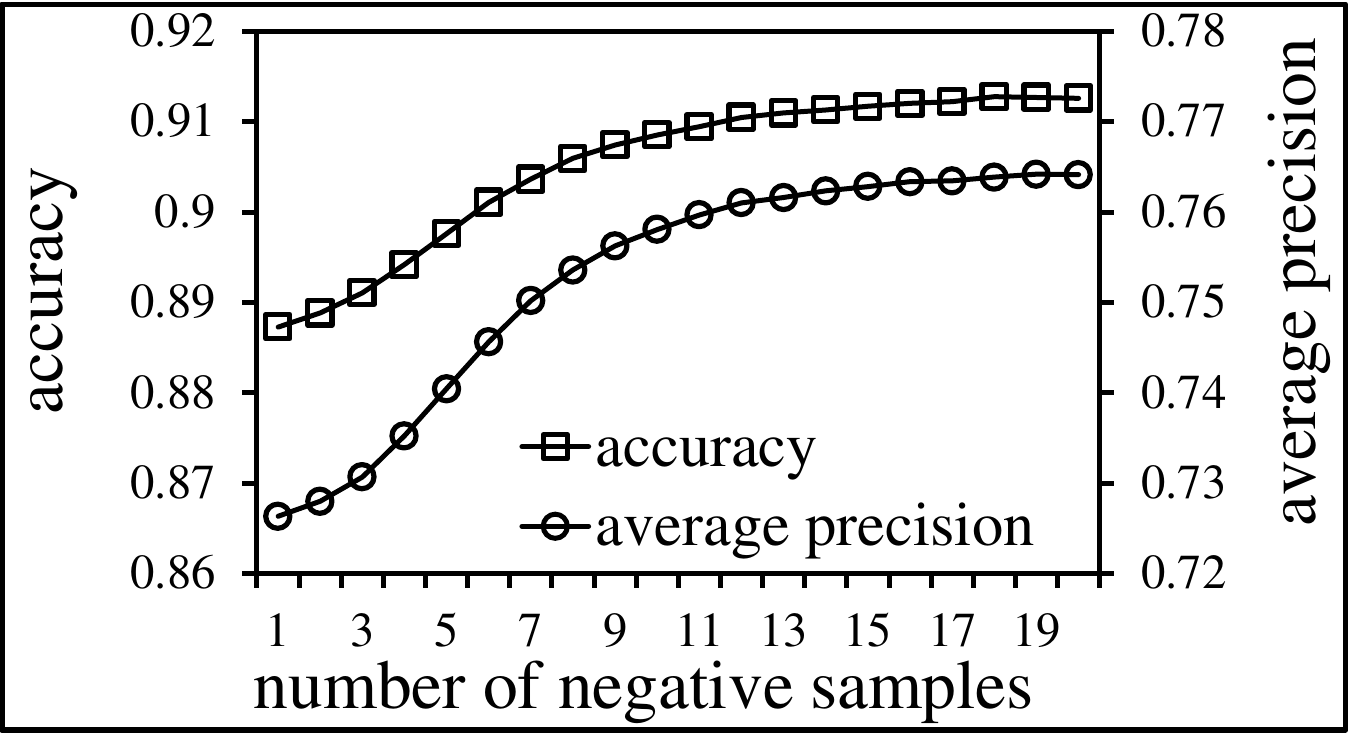}}
\hspace{0.05\textwidth}
\subfigure[Taxi data]{
\includegraphics[height=0.13\textheight]{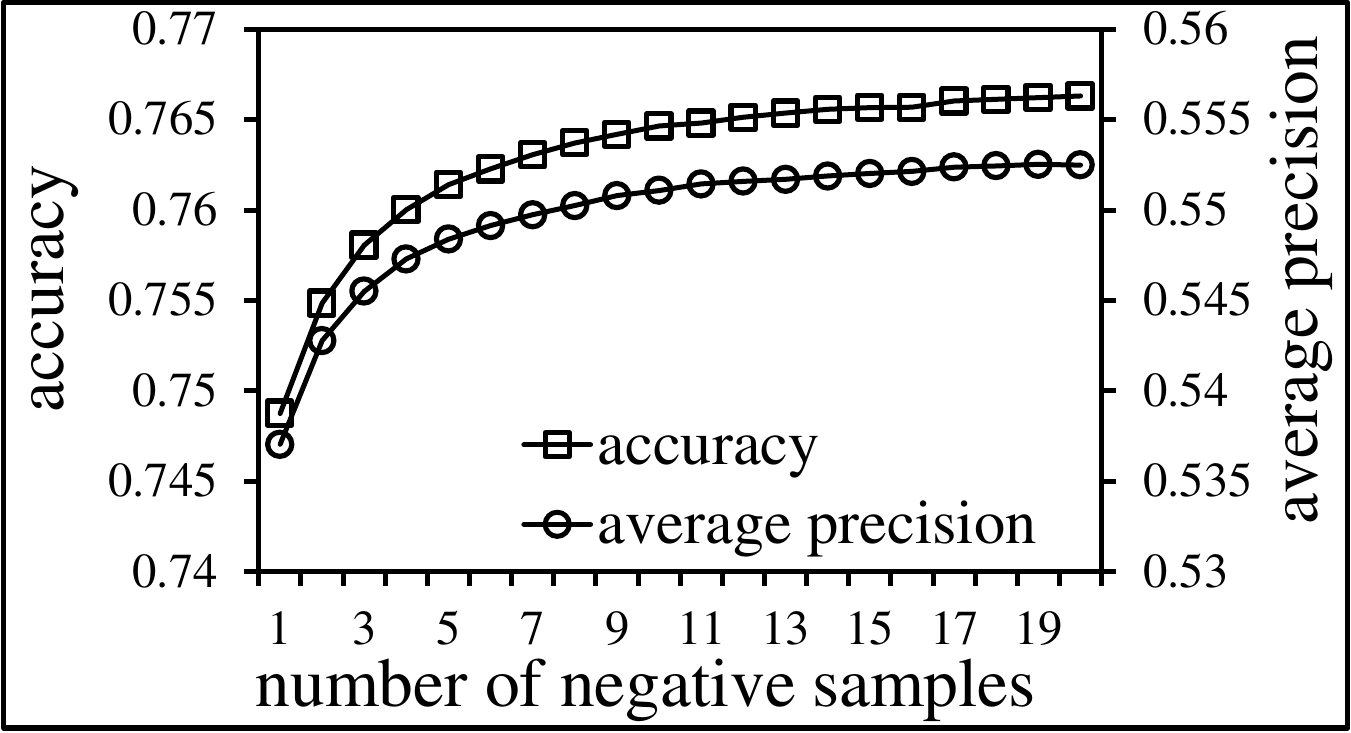}}
\caption{Effect of the number of negative samples $M$.}
\label{fig:neg}
\end{figure*}

Before applying MPE to our data, we need to map the time-stamp of each record to the time slot it belongs to. We set the size of slot at 1, 5, 10, 15, 30, 60 and 120 minutes respectively and evaluate the performances. The optimal size of the slot is 30 minutes for the VPR data, and 15 minutes for the Taxi data. Then we measure the effect of the parameters in MPE, including the regularization parameter $\lambda$, the number of vector dimensionality $D$, and the number of negative samples $M$, and tune them one by one on the validation set. The tuning results on both datasets with top-3 accuracy and average precision are reported in Fig.~\ref{fig:reg}, Fig.~\ref{fig:dim}, and Fig.~\ref{fig:neg}, and the impacts of varying these parameters are discussed below.

We first vary the regularization parameter $\lambda$ from $10^{-1}$ to $10^{-6}$, which could prevent over-fitting. As shown in Fig.~\ref{fig:reg}, the accuracy and average precision improve significantly when we decrease $\lambda$ from $10^{-1}$ to $10^{-3}$, and keep stable as we further decrease it.

Next we tune the embedding's dimensionality $D$ and the results are shown in Fig.~\ref{fig:dim}. We observe that on both datasets, the prediction performances improve as $D$ increases, and remain constant when $D$ is greater than 250. Finally, we tune the number of negative samples $M$ from 1 to 20 and report the results in Fig.~\ref{fig:neg}. The accuracy and average precision improve for both datasets as the number of negative samples $M$ increases, and vary little after $M=15$. Note that, it costs more time to complete training with the increase of $M$.

\subsection{Visualization of Embedding Vectors}
MPE embeds objects, locations and time slots in a low-dimensional latent space, which allows us to visually explore the relations among objects or time slots.

\subsubsection{Object visualization}
Since we all know that the movement patterns of taxis and private cars are quite different, we would like to see whether they can be visually distinguished in a projected space with MPE. We therefore randomly sample 2,000 taxis and 2,000 private cars from the VPR data, and obtain the embedding vectors of the corresponding objects. Fig.~\ref{fig:visualization}(a) shows a 2D t-SNE \cite{maaten2008visualizing} projection for these embedding vectors (taxis are colored green and private cars are yellow). Two obvious classes can be observed in the figure, explicitly proving that embedding vectors are effective features for object classification.

\subsubsection{Time visualization}
\begin{figure}[!t]
\centering
\subfigure[Object visualization (taxis are represented with green and private cars are with yellow.)]{
\includegraphics[width=0.4\textwidth]{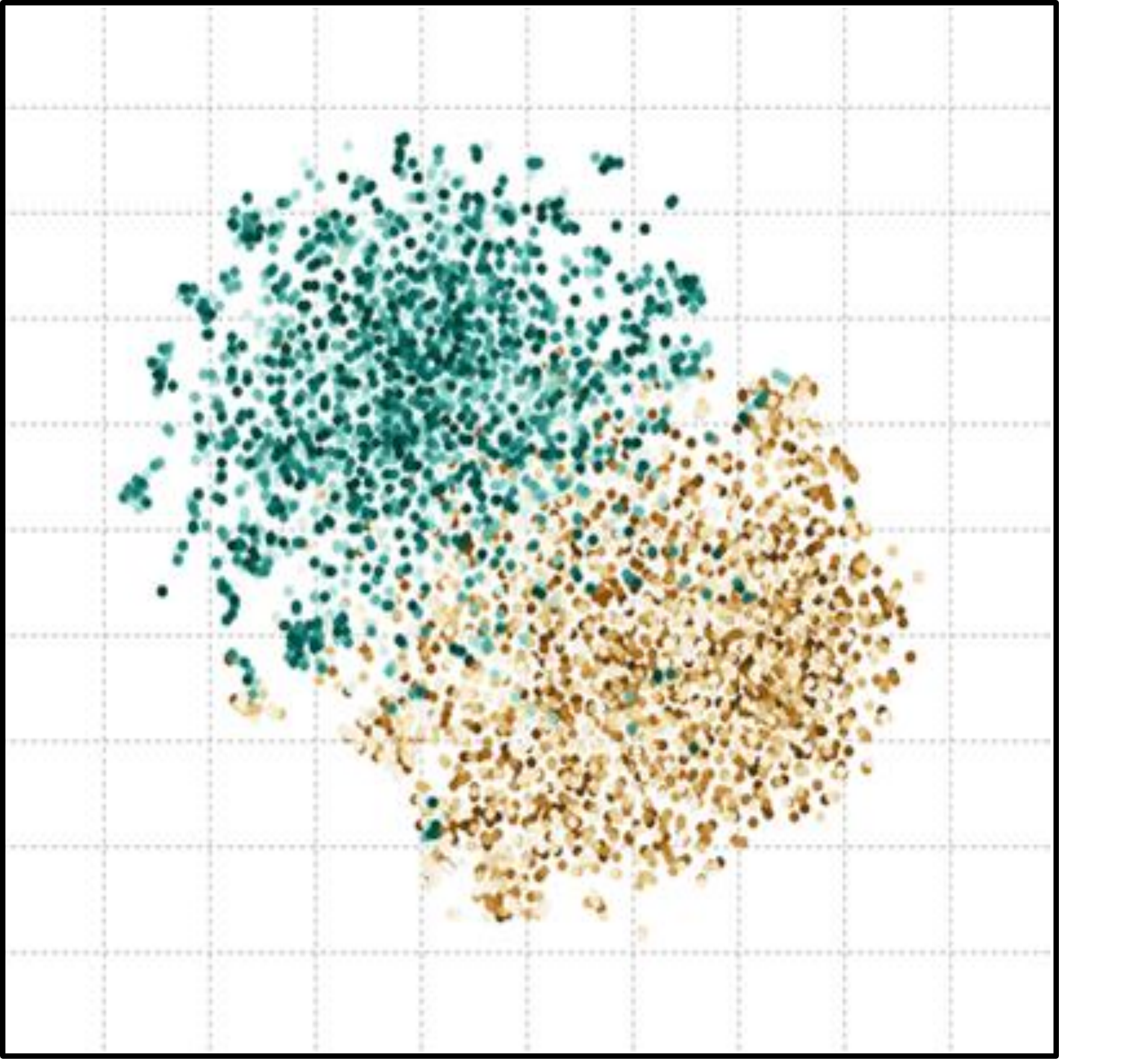}}
\hspace{0.05\textwidth}
\subfigure[Time visualization (96 time slots are represented with gradient colors.)]{
\includegraphics[width=0.4\textwidth]{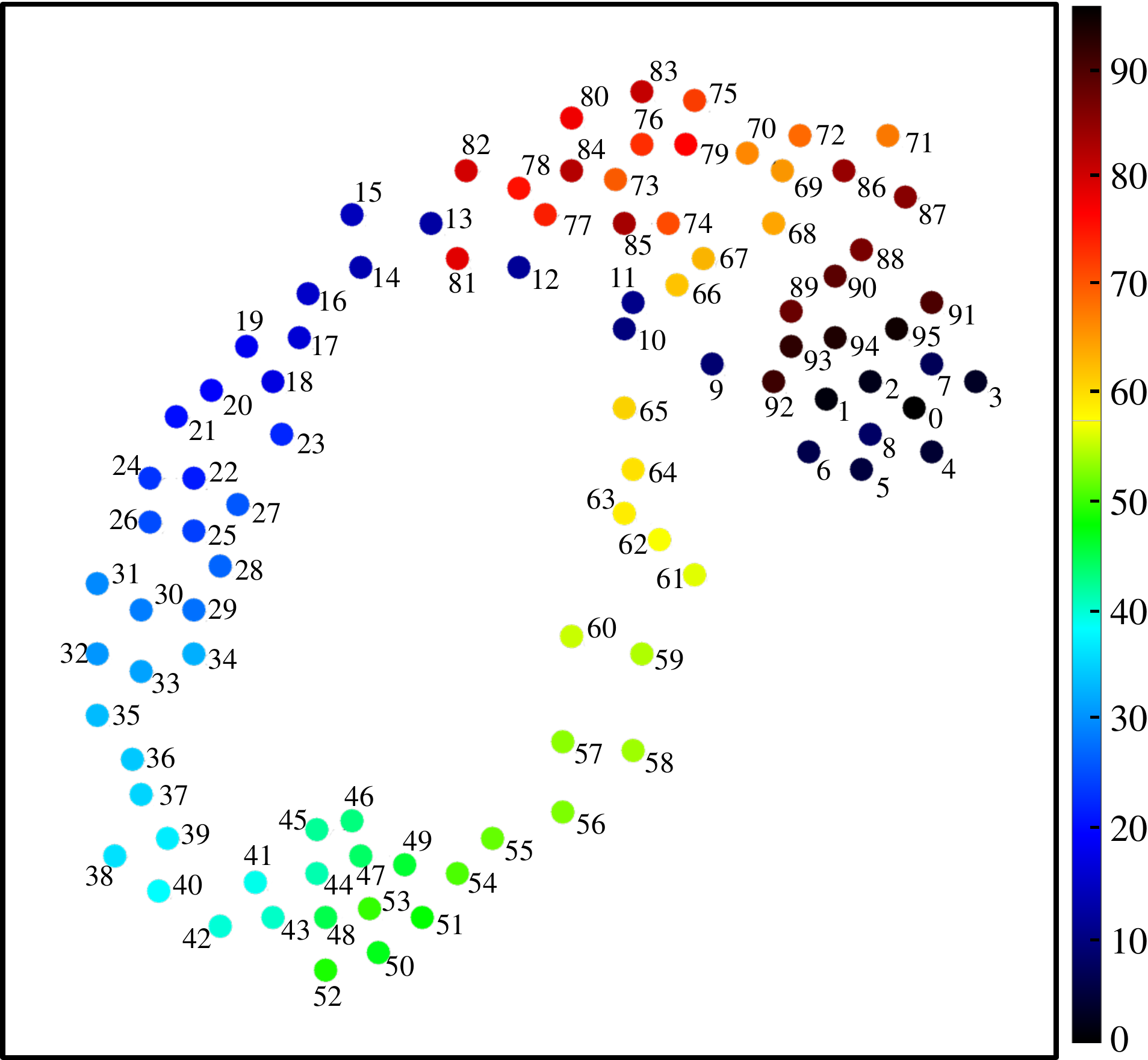}}
\caption{Visualization of embedding vectors.}
\label{fig:visualization}
\end{figure}

We use MPE to embed time slots in a latent space with the taxi trajectory data. Fig.~\ref{fig:visualization}(b) shows the 2D t-SNE projection for the vectors of 96 time slots (the size of slots is 15 minutes for the Taxi data based on the setting in Section~\ref{para}). We illustrate these time slots with gradient colors, and notice that: (1) the time slots roughly scatter in a ring form and the adjacent ones are still close to each other, which is in line with our common sense; (2) the distances between ``symmetrical'' time slots in the ring are pretty large. For example, the positions of slots in the morning are far from those in the afternoon. One likely reason is that there exist different functional regions in a city \cite{yuan2015discovering}, and vehicles often move in opposite directions in the morning and in the afternoon (e.g., going to work vs. going home); (3) the positions of time slots in the evening (i.e., the points in the upper right corner of Fig.~\ref{fig:visualization}(b)) are relatively more concentrated, as human mobility patterns are more random at night in contrast to showing clear tendencies during the day.

\subsection{Discussion}
Different from the traditional mobility pattern mining methods (e.g., Markov models, Bayes models), our proposed MPE sheds the light on modeling human movement patterns with the traffic trajectory data from a novel angle. With the distributed embedding vectors, we are able to not only predict next locations, but also compute the similarities between objects (or time slots) and visualize them, which cannot be achieved by the traditional methods.

Furthermore, we pay much attention to the traffic trajectory data in a real-world transportation system, which is different from the check-in data in the location-based social network. Specifically, people's driving trajectories are restricted by the road network, and the characteristic of ``phantom transitions'' does not exist; whereas people's visiting order of POIs is relatively random, as each POI represents an activity (e.g., eating in a restaurant, studying in a library) and no external restrictions are imposed on the moving patterns. Therefore, the existing embedding methods \cite{zhou2016general,feng2015personalized,zhao2017geo} modeling check-in data mainly focus on the correlation of POIs within a trajectory, for the task of POI recommendation; our proposed MPE devotes to modeling the transition of locations in a trajectory, for the task of next location prediction.

\section{Conclusion}\label{conclusion}
In this paper, considering the unique characteristics of traffic trajectory data, we have proposed a novel Mobility Pattern Embedding (MPE) method to learn human mobility patterns by jointly modeling sequential, personal, and temporal factors. Specifically, we project objects, time slots, current locations and next locations together as points in a low-dimensional latent space through MPE. Such embedding vectors could be exploited in many tasks, such as next location prediction and visualization. Finally, we evaluate the performances of MPE on two real datasets, and experimental results show that the proposed method outperforms state-of-the-art baselines significantly.



\end{document}